\documentclass{sbc2023}%

\usepackage{graphicx}
\usepackage[utf8]{inputenc}
\usepackage[misc,geometry]{ifsym} 
\usepackage{fontawesome}
\usepackage{orcidlink} 
\usepackage{float}

\usepackage{color}
\usepackage{hyperref} 
\usepackage{aas_macros}
\usepackage[bottom]{footmisc}
\usepackage{supertabular}
\usepackage{afterpage}
\usepackage{url}
\usepackage{pifont}
\usepackage{multicol}
\usepackage{tikz}
\usepackage{multirow}
\usepackage{amssymb}
\usepackage{amstext}
\usepackage{amsmath}
\usepackage{amsthm}
\usepackage{amsfonts}
\usepackage{booktabs}
\usepackage{multicol}
\usepackage{multirow}
\usepackage{graphicx}
\usepackage{url}
\usepackage{color}
\usepackage{mdframed}
\usepackage{lmodern}
\usepackage{xcolor}
\usepackage{listings}

\setcitestyle{square}

\definecolor{orcidlogo}{rgb}{0.37,0.48,0.13}
\definecolor{unilogo}{rgb}{0.16, 0.26, 0.58}
\definecolor{maillogo}{rgb}{0.58, 0.16, 0.26}
\definecolor{darkblue}{rgb}{0.0,0.0,0.0}
\hypersetup{colorlinks,breaklinks,
            linkcolor=darkblue,urlcolor=darkblue,
            anchorcolor=darkblue,citecolor=darkblue}

\title[BRoverbs - Measuring how much LLMs understand Portuguese proverbs]{BRoverbs - Measuring how much LLMs understand Portuguese proverbs}

\author[Almeida et al. 2025]{

\affil{\textbf{Thales Sales Almeida}~\orcidlink{0009-0006-9568-9331}~[~\textbf{Institute of Computing, University of Campinas. Maritaca AI}~|\href{mailto:thalesrogerio@gmail.com}{~\textbf{\textit{t224732@dac.unicamp.br}}}~]}

\affil{\textbf{Giovana Kerche Bonás}~\orcidlink{0009-0001-9460-8353}
[~\textbf{Institute of Computing, University of Campinas. Maritaca AI}~|\href{mailto:g216832@dac.unicamp.br}{~\textbf{\textit{g216832@dac.unicamp.br}}}~]}

\affil{\textbf{João Guilherme Alves Santos}~\orcidlink{0000-0001-5307-5338}[~\textbf{Institute of Computing, University of Campinas}~|\href{mailto:j199624@dac.unicamp.br}{~\textbf{\textit{j199624@dac.unicamp.br}}}~]}

}

\date{2025}

\begin{document}

\begin{frontmatter}
\maketitle

\begin{mail}
Institute of Computing, University of Campinas, Av. Albert Einstein, 1251, Campinas, SP, 13083-852, Brazil \\
Maritaca AI, Campinas, SP, Brazil
\end{mail}





\begin{abstract}
\textbf{Abstract.~}
\noindent Large Language Models (LLMs) exhibit significant performance variations depending on the linguistic and cultural context in which they are applied. This disparity signals the necessity of mature evaluation frameworks that can assess their capabilities in specific regional settings. In the case of Portuguese, existing evaluations remain limited, often relying on translated datasets that may not fully capture linguistic nuances or cultural references. Meanwhile, native Portuguese-language datasets predominantly focus on structured national exams or sentiment analysis of social media interactions, leaving gaps in evaluating broader linguistic understanding. To address this limitation, we introduce BRoverbs, a dataset specifically designed to assess LLM performance through Brazilian proverbs. Proverbs serve as a rich linguistic resource, encapsulating cultural wisdom, figurative expressions, and complex syntactic structures that challenge the model comprehension of regional expressions. BRoverbs aims to provide a new evaluation tool for Portuguese-language LLMs, contributing to advancing regionally informed benchmarking. The benchmark is available at \url{https://huggingface.co/datasets/Tropic-AI/BRoverbs}.
\end{abstract}


\end{frontmatter}








\section{Introduction}

In recent years, significant investments have been made in the development and expansion of large language models (LLMs). As these models become more integrated into real-world applications~\citep{li2023large, liang2024mapping, rane2023contribution}, evaluating their abilities across various tasks has become increasingly important. Reliable and varied evaluation methods help measure how well LLMs understand language, handle different tasks, and perform across languages and cultures.

Although LLMs have gained increasing global attention, research has predominantly focused on English, leaving evaluation datasets in other languages, such as Portuguese, relatively scarce~\citep{giagkou2023europeanlanguages}. As a result, many studies on Portuguese LLMs rely, in part, on translating existing tasks to measure performance~\citep{almeida2024sabi, correa2024tucano, larcher2023cabrita}. These translated tasks often fail to capture regional and cultural nuances. This issue was highlighted by Sabiá~\citep{almeida2024sabi}, which observed that after pretraining on Portuguese data, the model showed greater performance gains on tasks originally designed in Portuguese than on translated ones.

Given the increasing global expansion of LLM applications~\citep{naveed2023comprehensive, zhao2023survey}, it is crucial to assess whether they effectively fulfill regional capabilities such as linguistic comprehension, cultural adaptability, and performance in native language tasks.


Recent studies have shown that LLMs perform unevenly across regions, particularly when dealing with underrepresented countries and contexts~\citep{myung2024blend, moayeri2024worldbench, almeida2025tiebe}. These disparities underscore the importance of native benchmarks that reflect local knowledge and cultural specificity.

The performance gap observed in benchmarks likely reflects the low representation of certain regions in training data. Studies on data governance have revealed a significant imbalance in data sources, with Latin American representation remaining exceedingly low and showing no signs of increasing. \citep{longpre2024bridging} audited training datasets over recent years, showing that data representation is overwhelmingly concentrated in North America, Western Europe, and China, while Latin America contributes less than 0.5\% of the total. Such audits motivate targeted evaluation—not because a single benchmark can resolve systemic bias, but because it can reveal where that bias manifests in model behavior.

This imbalance not only impacts model performance but also reinforces systemic biases, further marginalizing non-dominant languages and cultural expressions in AI-driven technologies. To better understand how such imbalances manifest, we propose BRoverbs, a dataset designed to evaluate the comprehension of Brazilian proverbs by LLMs. Proverbs are deeply rooted in cultural and linguistic contexts, making them a suitable signal of a model's ability to grasp regional semantics and idiomatic expressions. Our benchmark represents a step forward in evaluating Brazilian regional knowledge.
 
Additionally, the creation of culturally grounded datasets like BRoverbs serves a social purpose. By concentrating on commonly shared folk wisdom encoded in Brazilian proverbs, we underscore the need for inclusive large language models. Such models should be capable of understanding not only the direct meaning of words but also the deeply woven cultural, historical, and linguistic nuances that shape diverse communities. This approach encourages equitable access to advanced language technologies for Brazilian Portuguese speakers.

Our contributions are as follows.

\begin{itemize}
\item We introduce BRoverbs, a new benchmark designed to evaluate LLMs understanding of Brazilian proverbs, improving Brazil's scenario of culturally aware LLM evaluation.
\item We evaluate a diverse set of LLMs, including commercial and open-source models of varying sizes, highlighting disparities in performance and the impact of model scale and pretraining data composition.
\item We discuss the implications of cultural underrepresentation in LLM training data and emphasize the importance of developing native benchmarks to ensure fair and accurate assessments of linguistic understanding in underrepresented languages.
\end{itemize}

\section{Related work}

The evaluation of large language models (LLMs) has been an active area of research, with numerous benchmarks designed to assess different aspects of model performance. These benchmarks focus on a wide variety of tasks, such as reading comprehension~\citep{dua2019drop, lai2017race, rajpurkar2016squad, kwiatkowski2019natural, yu2020reclor}, sentimental analysis and subjectivity~\citep{potts2020dynasent, alfina2017hate}, reasoning and factual knowledge retrieval~\citep{yu2020reclor, thorne2018fever, jiang2020x, ho2020constructing}, generation and creativity~\citep{srivastava2022beyond, chen2021evaluating, hasan2021xl}, and others~\citep{rudinger2018gender, parrish2021bbq, wang2019superglue}. While these benchmarks provide a broad foundation for evaluating LLMs, they predominantly focus on English, leaving gaps in assessing model performance in other languages. In Portuguese, evaluation resources remain scarce, often requiring the translation of existing English-language tasks, which can introduce inconsistencies and fail to capture the linguistic and cultural nuances necessary for robust assessments.

\subsection{Evaluating LLMs in Portuguese and Iberian Languages}: Trends and Gaps

Recent efforts to evaluate LLMs in Portuguese have followed two main trends. The first trend focuses on structured exams used for human evaluation, such as benchmarks based on the Brazilian national university entrance exam (ENEM)~\citep{silveira2017university} or other universities entrance exams~\citep{almeida2023bluex} and based on the Order of Attorneys of Brazil national bar exam~\citep{delfino2017passing}. While the second trend, centers around social media classifications, such as sentimental analysis in tweets proposed by TweetSentBR~\citep{brum2017building}, social media hate detection through datasets like HateBR\citep{vargas2021hatebr} and PT Hate Speech\citep{fortuna2019hierarchically}, and other studies targeting the analysis of informal, user-generated content, such as fake news detection or similar tasks. 

While both approaches offer valuable insights into model performance, they fail to fully capture key aspects of factual recall, cultural adaptation, and regional linguistic variation. Notable exceptions include Faquad~\citep{sayama2019faquad}, an small extractive QA dataset, and TiEBe~\citep{almeida2025tiebe}, which focuses on open-domain QA about Brazilian events. However, these efforts remain limited in scope, highlighting the need for benchmarks that evaluate LLMs within richer cultural and linguistic contexts. A recent contribution in this direction is IberoBench ~\citep{baucells2025iberobench} which targets the evaluation gap for Iberian languages—including Basque, Catalan, Galician, European Spanish, and European Portuguese. The benchmark comprises 62 tasks grouped into 179 subtasks across ten skill categories (e.g., commonsense reasoning, NLI, QA, summarization, translation) and assesses 33 LLMs in zero-shot and few-shot settings, revealing that performance in these languages still trails state-of-the-art results in English.

\subsection{Regional Disparities in LLM Evaluation}

A recent wave of research has introduced benchmarks designed to address biases and regional disparities in LLM factual recall. Notable among them are WorldBench~\citep{moayeri2024worldbench}, TiEBe~\citep{almeida2025tiebe} and BLEnD~\citep{myung2024blend}. These benchmarks reveal systemic biases in LLMs and highlight areas where model performance remains uneven. 

WorldBench~\citep{moayeri2024worldbench} is designed to assess geographic disparities in factual recall using data per country from the World Bank, an international financial institution that provides funding, advice and research to help countries reduce poverty and promote sustainable development. The benchmark has revealed significant biases in LLM performance based on region and income level, showing that error rates are notably higher for lower-income countries, particularly in regions such as Sub-Saharan Africa. These disparities emphasize the need for more representative and globally inclusive evaluation methodologies.

Similarly, TiEBe~\citep{almeida2025tiebe} expands LLM evaluation by addressing real-world knowledge and historical events with a dataset of over 11,000 question-answer pairs. On Wikipedia, there are annual pages that mention significant events both globally and in various countries. These pages list notable events, each accompanied by a set of external citation links, often pointing to journalistic sources that provide further context. TiEBe utilizes this information, spanning a 10-year interval, to generate question-answer pairs about the events. The purpose is to assess whether a language model can recall information from the original news sources, even though the model will not have direct access to these documents during testing. This approach ensures that the questions reflect major events and figures in a consistent and structured manner. It also reveals that LLMs often fail to represent events consistently across different regions, including Portuguese-speaking countries such as Brazil and Portugal.

BLEnD~\citep{myung2024blend} takes a different approach by evaluating LLMs' understanding of daily knowledge across multiple cultural and linguistic contexts. It includes 52.6k manually crafted question-answer pairs covering 16 countries and 13 languages, revealing that LLMs perform better in cultures and languages with stronger digital representation. While models generate more accurate responses in native languages for mid-to-high-resource languages, they often provide more reliable answers in English when dealing with low-resource languages. This reflects the ongoing challenge of training LLMs to effectively represent diverse linguistic and cultural knowledge.

Collectively, WorldBench, TiEBe, and BLEnD underscore the pressing need for culturally aware evaluation benchmarks that extend beyond mainstream datasets. These efforts expose significant gaps in LLMs' regional knowledge and cultural comprehension, highlighting the broader challenge of ensuring AI systems equitably serve diverse global communities. At the same time, they demonstrate the potential for further research in this area, paving the way for new methodologies that assess how well models handle complex cultural expressions.

\subsection{LLM Performance in Different Languages}

While previous studies focused on regional disparities in factual recall, a growing body of work has turned its attention to a different but related challenge: the performance of LLMs across diverse languages, particularly those with limited digital representation. Linguistic diversity introduces unique obstacles, as models often struggle with languages that lack large-scale training data, affecting their ability to generate accurate and contextually appropriate responses.

Recent benchmarks have highlighted the limitations of LLMs in capturing region-specific knowledge, particularly in languages with limited digital presence. For example, Pariksha~\citep{watts2024pariksha} evaluates 30 LLM models in 10 Indic languages, conducting a large-scale comparison between human and LLM-based evaluations. The project reveals that discrepancies in evaluation arise in lower-resource languages such as Bengali and Odia, where human and machine evaluations diverge significantly, reinforcing concerns about how models interpret and assess culturally embedded concepts.

These studies reinforce the idea that cultural and linguistic biases in LLMs are a widespread issue, not limited to Portuguese. They also illustrate how regionally adapted benchmarks can help uncover deficiencies in the ability of LLMs to generalize knowledge between different cultures. 

\subsection{Figurative-Language Benchmarks}

Figurative language—idioms, metaphors, similes and, above all, proverbs—conveys meaning that is not compositionally recoverable from the literal sense of its words.  Because large language models (LLMs) are typically trained to predict text token-by-token, their ability to recognize and interpret such nonliteral signals is a stringent test of genuine contextual understanding. Dedicated benchmarks have emerged to probe this ability.

ProverbEval~\citep{azime2024proverbeval} contributes to the evaluation of the understanding of proverbs by LLMs, particularly in the context of low-resource languages. It introduces a multilingual benchmark to evaluate LLMs’ understanding of proverbs, with a particular emphasis on low-resource and typologically diverse languages, presenting evaluations in Amharic, Tigrinya, Ge’ez, Afaan Oromo, and English.

The benchmark is structured in three different types of tasks: (1) a meaning-based multiple-choice task, in which models select the correct interpretation of a proverb from four detailed options; (2) a fill-in-the-blank task, which tests the model’s ability to recognize conventional phrasings in known proverbs; and (3) a generation task, where models must retrieve the appropriate proverb given a detailed description of its meaning or situational use. This setup allows researchers to examine not only a model's figurative-language understanding but also its ability to transfer proverb knowledge across languages and scripts.

While this project shares similarities with BRoverbs, particularly in its focus on assessing regional linguistic comprehension through culturally embedded expressions, it adopts a different methodological approach. ProverbEval's approach sheds light on the key challenges in evaluating how well models grasp proverb meanings across different linguistic landscapes. In contrast, BRoverbs provides a focused analysis of Brazilian Portuguese, offering a detailed examination of the regional linguistic complexities unique to this language.

Complementary to these proverb-focused datasets, ~\citep{mi2024rolling} present Rolling the DICE on Idiomaticity, a controlled contrastive benchmark that pairs 402 English idioms with twin sentences—one literal, one figurative—differing only in their surrounding context.  Results on 13 popular LLMs (GPT-4o, Llama-3, etc.) show that even the strongest models struggle to resolve the literal-versus-figurative ambiguity, often defaulting to the figurative sense and achieving below-perfect “strict consistency” across sentence pairs.  This finding highlights that figurative interpretation remains challenging even in high-resource languages and supports the need for culturally grounded evaluations like BRoverbs.

\section{Methodology}
\begin{figure}[ht]
\centering
\centering
    \includegraphics[width=0.9\linewidth]{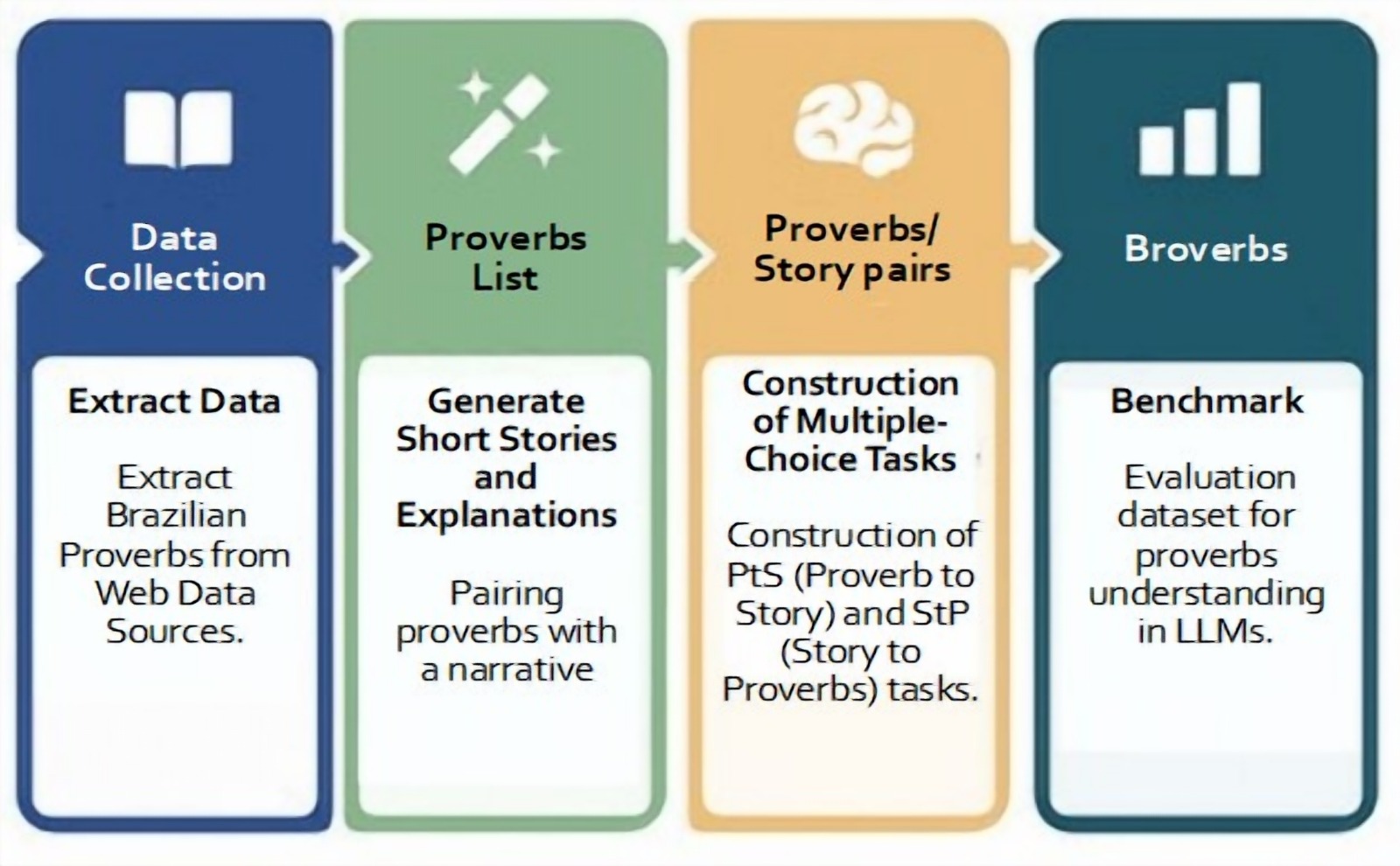}
    \caption{Methodological flow of \textit{BRoverbs}, illustrating the steps of data collection, story generation, creation of questions and answers, and model evaluation.}
    \label{fig:pipeline}
\end{figure}

In this section, we describe the pipeline for creating BRoverbs, which is illustrated in Figure~\ref{fig:pipeline}.  Our goal was to build a dataset that connects popular Brazilian proverbs to short narratives and propose tasks that assess how effectively language models can correctly associate each proverb with its corresponding story and vice versa.

\subsection{Data Collection and Story Generation}
Our process began by collecting Brazilian proverbs from online sources. We conducted three web searches~\footnote{As buscas foram: 'lista de proverbios brasileiros', 'ditados populares brasileiros', 'proverbios populares brasileiros'} for Brazilian proverbs and gathered the top five results for each query, resulting in 15 different sources. Proverbs were extracted from all sites, then grouped using lexical clustering based on fuzzy string matching. Duplicates were then manually removed, yielding a final set of 196 unique proverbs.

For each proverb, we used GPT-4 to generate three short narratives illustrating its meaning, without quoting or referencing the proverb directly. The prompt also asked GPT-4 to provide a brief explanation of the moral or practical meaning behind each proverb for verification purposes. The used prompt is available at appendix~\ref{sec:prompts}.

Each of the three stories for each proverb was manually verified, and some adjustments were made when the meaning of the story did not correctly reflect the meaning of the expression, and in some rare cases, the stories were completely rewritten. In this process, we also dropped three proverbs completely, such as "Batatinha quando nasce, se esparrama pelo chão", due to not reaching a consensus for their meaning.

Thus, we obtained:
\begin{itemize}
    \item 579 short stories (3 for each of the 193 proverbs left from manual verification).
    \item A concise explanatory summary for each proverb.
\end{itemize}

\subsubsection{Distribution of Story Length in Characters}
\label{subsec:distribution-story-length}

\begin{figure}[h!]
    \centering
    \includegraphics[width=0.9\linewidth]{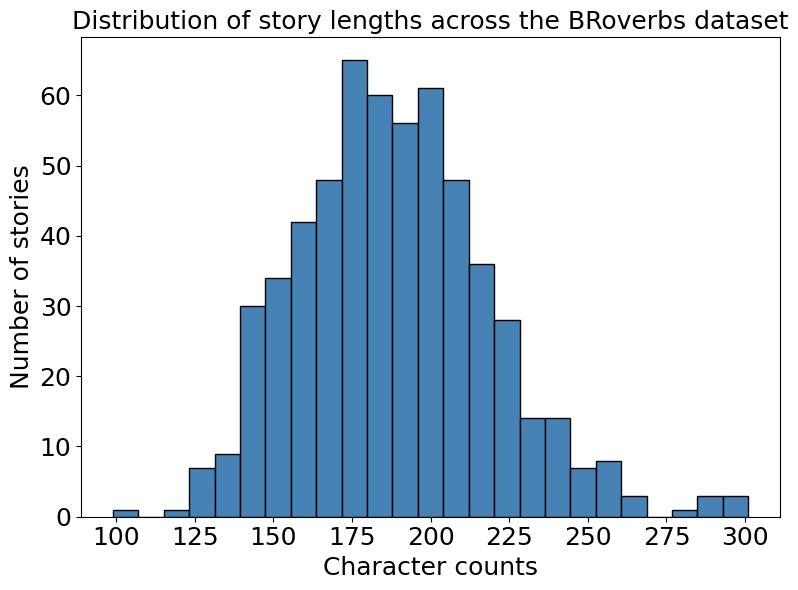}
    \caption{Distribution of story lengths (in characters) across the BRoverbs dataset.}
    \label{fig:distribuicao-historias}
\end{figure}

To gain deeper insights into the textual complexity of the 579 stories, we computed the length of each story by counting the total number of characters. Figure~\ref{fig:distribuicao-historias} shows the histogram of character counts (x-axis) against the number of stories (y-axis). We observe that most stories lie within a similar range of lengths, reflecting a relatively consistent output from the generative prompt.

\subsubsection{Jaccard Distance Across Triplet Stories}
\label{subsec:jaccard-distance}
\begin{figure}[h!]
    \centering
    \includegraphics[width=0.9\linewidth]{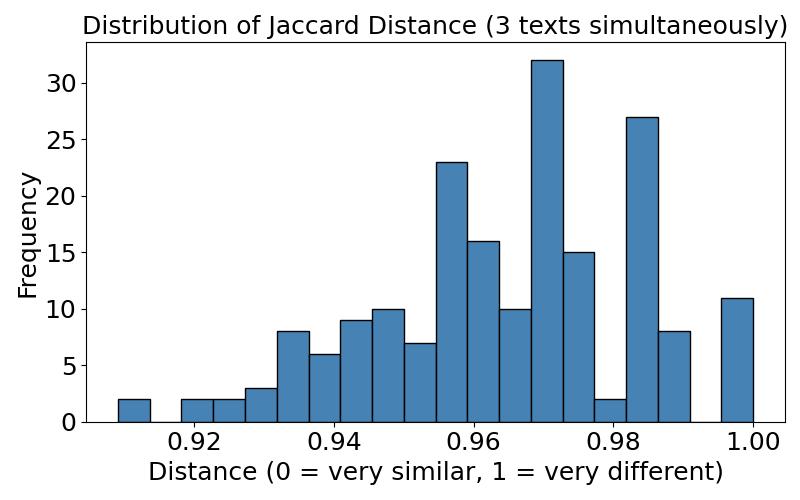}
    \caption{Distribution of Jaccard distances among the three short stories for each proverb.}
    \label{fig:jaccard-distances}
\end{figure}

In addition to examining story length, we also measured how distinct the three short stories generated for each proverb were from one another, using the Jaccard distance. As shown in Figure~\ref{fig:jaccard-distances}, these distances are generally quite high—most values exceed 0.9. This indicates that the short stories typically share relatively few words in common, reflecting a high degree of lexical diversity. Such diversity is important to ensure each story presents a unique illustration of the same underlying proverb without simply repeating the same wording.

\subsection{Generation of Question-Answer Pairs}
After compiling the proverbs and their corresponding narratives, we designed two tasks to assess whether language models can accurately match proverbs to stories:
\begin{itemize}
    \item PtS (Proverb to Story): Given a specific proverb, the model must identify which of the five provided short stories best represents that proverb. One of the short stories matches the proverb accurately, while the remaining four options were randomly selected from the full set of 579 stories. An example of this task can be seen in Figure ~\ref{fig:example-broverbs-PtS}.
    \item StP (Story to Proverb): Given a short story, the model must select which of the five presented proverbs is most closely related to that story. Here, too, one proverb was correct, and the remaining four were randomly chosen from the total pool of proverbs. An example of this task can be seen in Figure ~\ref{fig:example-broverbs-StP}.
\end{itemize}

Because some proverbs have similar meanings—for example, "Quem com ferro fere, com ferro será ferido" and "Tudo que vai, volta"—randomly selecting stories and proverbs for the alternatives can result in more than one correct answer. To address this, we manually reviewed the distractor options to ensure that none of the three randomly chosen proverbs unintentionally matched the meaning of the correct answer, making corrections whenever such overlaps were found.

This process was performed in two stages. First, we reviewed each question in the dataset to ensure the alternatives were appropriate. Then,  after running four commercial models (GPT-4o, GPT-4o-mini, Sabiazinho-3, and Sabiá-3), we re-examined all questions that were answered incorrectly by at least one of the models—this accounted for about 12\% of the total questions. From these questions, around 35\% were modified.

\begin{figure*}[t]
    \centering
    \begin{mdframed}[backgroundcolor=gray!10, roundcorner=10pt]
    \textbf{Qual das histórias melhor corresponde ao provérbio dado?}

    \vspace{0.5em}
    \textbf{Provérbio:} O que vem fácil, vai fácil

    \vspace{0.5em}
    \begin{enumerate}[label=\textbf{\Alph*)}]
        \item Clara estava animada com a ideia de estudar em uma universidade no exterior. No entanto, ao chegar lá, enfrentou saudades de casa e dificuldades com o idioma.
        \item No campeonato de xadrez, Miguel venceu a maioria dos jogos com facilidade devido à sua experiência. Seu adversário, Tiago, teve que estudar cada movimento meticulosamente para conseguir alcançar a próxima fase.
        \item Tatiane estava empolgada com sua nova coleção de sapatos. No entanto, ao passar por uma vitrine, viu um par que ainda não tinha e não conseguiu resistir à tentação de comprá-lo.
        \item \textbf{Carlos ganhou uma grande quantia em um jogo de azar e decidiu gastar tudo em festas e viagens. Em poucos meses, estava sem dinheiro e com dívidas.}
        \item Durante uma reunião de família, Carlos fez um comentário sarcástico sobre a comida preparada por sua tia. Ela, sentida, respondeu de maneira ríspida, lembrando-lhe de um antigo erro embaraçoso.
    \end{enumerate}
    
    \end{mdframed}
    
    \vspace{1em}

    \begin{mdframed}[backgroundcolor=gray!10, roundcorner=10pt]
    \textbf{Which of the stories best fits the given proverb?}

    \vspace{0.5em}
    \textbf{Provérbio:} What comes easy, goes easy.

    \vspace{0.5em}
    \begin{enumerate}[label=\textbf{\Alph*)}]
        \item Clara was excited about the idea of studying at a university abroad. However, upon arriving there, she faced homesickness and difficulties with the language.
        \item In the chess championship, Miguel won most of the games with ease due to his experience. His opponent, Tiago, had to study each move meticulously to reach the next round.
        \item Tatiane was excited about her new shoe collection. However, when passing by a shop window, she saw a pair she didn't have yet and couldn't resist the temptation to buy it.
        \item \textbf{Carlos won a large amount of money in a game of chance and decided to spend it all on parties and trips. Within a few months, he was out of money and in debt.}
        \item During a family meeting, Carlos made a sarcastic comment about the food prepared by his aunt. She, hurt, responded sharply, reminding him of an old embarrassing mistake.
    \end{enumerate}

    \end{mdframed}
    
    \caption{Example from the \textit{BRoverbs} dataset: The first section shows a \textbf{PtS (Proverb to Story)} task in Brazilian Portuguese, and the second presents the \textbf{literal} English translation of the same task.}

    \label{fig:example-broverbs-PtS}
\end{figure*}

\begin{figure*}[t]
    \centering
    \begin{mdframed}[backgroundcolor=gray!10, roundcorner=10pt]
    \textbf{Qual dos provérbios melhor corresponde à história dada?}

    \vspace{0.5em}
    \textbf{História:} Mariana viu um homem com roupas rasgadas sentado na praça e logo pensou que ele era desleixado. Mais tarde, descobriu que ele era um renomado artista de rua, famoso por sua generosidade.

    \vspace{0.5em}
    \begin{enumerate}[label=\textbf{\Alph*)}]
        \item Em time que está ganhando, não se mexe
        \item Um homem prevenido vale por dois
        \item \textbf{Não julgue um livro pela capa}
        \item Todos os caminhos levam à Roma
        \item Cada um sabe onde o sapato aperta
    \end{enumerate}

    \end{mdframed}
    
    \vspace{1em}

    \begin{mdframed}[backgroundcolor=gray!10, roundcorner=10pt]
    \textbf{Which proverb best matches the given story?}

    \vspace{0.5em}
    \textbf{Story:} Mariana saw a man in torn clothes sitting in the square and immediately thought he was careless. Later, she discovered that he was a renowned street artist, famous for his generosity.

    \vspace{0.5em}
    \begin{enumerate}[label=\textbf{\Alph*)}]
        \item In a team that is winning, don't make changes.
        \item A forewarned man is worth two.
        \item \textbf{Don't judge a book by its cover.}
        \item All roads lead to Rome.
        \item Each one knows where the shoe pinches.
    \end{enumerate}

    \end{mdframed}
    
    \caption{Example from the \textit{BRoverbs} dataset: The first section shows a \textbf{StP (story to proverb)} task in Brazilian Portuguese, and the second presents the \textbf{literal} English translation of the same task.}

    \label{fig:example-broverbs-StP}
\end{figure*}

\subsection{Model Evaluation}

We evaluated these tasks using a range of language models, including both open-source and commercial ones. Each model was tested on its ability to correctly align a proverb with its corresponding short story (PtS task) and vice versa (StP task). The performance outcomes served as a measure of how effectively each model could interpret and map the semantic content of the proverbs to the narrative situations generated for them. We used an RTX A6000 for all our experiments with open-source models.
All evaluations were conducted in a 1-fewshot scenario. We randomly selected a different example from the dataset for each test instance to serve as the single-shot illustration, ensuring it was not the same as the input being evaluated, preventing overlap and avoiding potential data leakage.

\begin{table*}[]
\centering
\setlength{\tabcolsep}{4mm}
\
\caption{Results for various LLMs in both BRoverbs tasks. The table also shows the model size and size of pretrain for open source models. }
\label{tab:main_results}
\begin{tabular}{@{}lccccc@{}}
\multicolumn{1}{l|}{Model}         & Model size & \multicolumn{1}{c|}{Pretrain size} & PtS  & StP  & Average \\ \midrule
\multicolumn{6}{c}{Large Commercial Models}                                                    \\ \midrule
\multicolumn{1}{l|}{GPT-4o~\citep{hurst2024gpto}}       & -    & \multicolumn{1}{c|}{-}        & 0.97 & 0.98 & 0.97 \\
\multicolumn{1}{l|}{Claude-3.5-sonnet~\citep{anthropic2024claude35sonnet}}      & -    & \multicolumn{1}{c|}{-}        & 0.96 & 0.98 & 0.97 \\ 
\multicolumn{1}{l|}{Sabiá-3~\citep{abonizio2024sabia3}}      & -    & \multicolumn{1}{c|}{-}        & 0.96 & 0.96 & 0.96 \\ \midrule
\multicolumn{6}{c}{Small Commercial Models}                                                   \\ \midrule
\multicolumn{1}{l|}{GPT-4o-mini~\citep{hurst2024gpto}}  & -    & \multicolumn{1}{c|}{-}        & 0.94 & 0.97 & 0.95 \\
\multicolumn{1}{l|}{Claude-3.5-haiku~\citep{anthropic2024claude35haiku}}  & -    & \multicolumn{1}{c|}{-}        & 0.95 & 0.94 & 0.95 \\
\multicolumn{1}{l|}{Sabiazinho-3~\citep{abonizio2024sabia3}} & -    & \multicolumn{1}{c|}{-}        & 0.94 & 0.96 & 0.95 \\ \midrule
\multicolumn{6}{c}{Medium open source models}                                                 \\ \midrule
\multicolumn{1}{l|}{Qwen 2.5 14b~\citep{yang2024qwen25}} & 14B  & \multicolumn{1}{c|}{18T}      & 0.92 & 0.94 & 0.93 \\
\multicolumn{1}{l|}{Qwen 2.5 7B~\citep{yang2024qwen25}}  & 7B   & \multicolumn{1}{c|}{18T}      & 0.83 & 0.83 & 0.83 \\
\multicolumn{1}{l|}{Llama3-8B~\citep{dubey2024llama3}}    & 8B   & \multicolumn{1}{c|}{15T}      & 0.48 & 0.75 & 0.61 \\
\multicolumn{1}{l|}{Llama2-7B~\citep{touvron2023llama2}}    & 7B   & \multicolumn{1}{c|}{2T}       & 0.22 & 0.25 & 0.23 \\
\multicolumn{1}{l|}{Llama1-7B~\citep{touvron2023llama}}    & 7B   & \multicolumn{1}{c|}{1T}       & 0.21 & 0.24 & 0.22 \\
\multicolumn{1}{l|}{Sabiá-7B~\citep{sabia}}     & 7B   & \multicolumn{1}{c|}{1T + 10B} & 0.23 & 0.33 & 0.28 \\ \midrule
\multicolumn{6}{c}{Small Open source models}                                                  \\ \midrule
\multicolumn{1}{l|}{Qwen 2.5 3B~\citep{yang2024qwen25}}  & 3B   & \multicolumn{1}{c|}{18T}      & 0.51 & 0.68 & 0.59 \\
\multicolumn{1}{l|}{Qwen 2.5 1.5B~\citep{yang2024qwen25}} & 1.5B       & \multicolumn{1}{c|}{18T}           & 0.41 & 0.57 & 0.49    \\
\multicolumn{1}{l|}{Tucano 2.4B~\citep{correa2024tucano}}  & 2.4B & \multicolumn{1}{c|}{500B}     & 0.19 & 0.20 & 0.19 \\
\multicolumn{1}{l|}{Tucano 1.1B~\citep{correa2024tucano}}  & 1.1B & \multicolumn{1}{c|}{250B}     & 0.19 & 0.20 & 0.19 \\
\multicolumn{1}{l|}{TinyLlama 3T~\citep{zhang2024tinyllama}} & 1.1B & \multicolumn{1}{c|}{3T}       & 0.19 & 0.20 & 0.19 \\
\multicolumn{1}{l|}{TinyLlama 1T~\citep{zhang2024tinyllama}} & 1.1B & \multicolumn{1}{c|}{1T}       & 0.18 & 0.22 & 0.20 \\
\multicolumn{1}{l|}{Curió 1.1B~\citep{building_curio}} & 1.1B & \multicolumn{1}{c|}{1T + 150B}       & 0.20 & 0.20 & 0.20
\end{tabular}
\end{table*}

\section{Results}

We now present the performance of various LLMs on the BRoverbs benchmark, analyzing how model size, training data, and language exposure affect task success.

\subsection{Model Results}
We evaluated our dataset using a diverse selection of LLMs, spanning both open-source and commercial models accessible via API. Among the commercial models, we included OpenAI's GPT-4o and GPT-4o-mini, Anthropic's Claude 3.5 Sonnet and Haiku, as well as Maritaca AI's Sabiá-3 and Sabiazinho-3, which represent Brazilian commercial alternatives. On the open-source side, we tested the Qwen 2.5B family~\citep{yang2024qwen25}, ranging from 1.5B to 14B parameters, a state-of-the-art option, along with three generations of LLaMA models~\citep{touvron2023llama,touvron2023llama2,dubey2024llama3} at approximately 7B parameters each. We also evaluated Sabiá 7B~\citep{sabia}, a further pretrained Portuguese variant of LLaMA-1 7B. To assess smaller models, we included Tucano 1.1B and 2.4B~\citep{correa2024tucano}, a recent family trained from scratch in Portuguese. Additionally, we considered TinyLlama 1T and 3T, 1.1B-parameter models trained on a large corpus of English data. Finally, we tested Curió-1.1B, which builds on TinyLlama 1T with an additional 150B tokens of Portuguese pretraining.

We separate the commercial models into two categories, "Large" and "Small", based on their price range. As for the open source models, we call "Medium" models above 7B parameters and "small" models with less than 7B parameters. The chosen models were selected to study how models with different sizes and training sizes perform in our tasks; the full results on both BRoverbs tasks, models sizes and pretrain sizes of the models are shown in Table~\ref{tab:main_results}, pretrain sizes in the format "X+Y" indicate a model under continued pretraining, where X is the number of tokens used in the base model, and Y the number of tokens used in the second pretrain.

First, we can see that both the 'Small' and 'Large' commercial models perform well in the task, all achieving an average between the two tasks superior to 95\%. Here is worth noting that while GPT-4o was used to help in story elaboration, it does not show a significant performance gap when compared to the other commercial models of similar cost.

One notable trend across almost all models is that the StP variant of the task is generally easier than the PtS task. This can be observed in the consistently higher scores for StP, even in models that perform poorly overall. The discrepancy between the two tasks suggests that identifying a proverb given a story is a simpler task than recognizing a story that matches the meaning of a given proverb. This is likely because many proverbs are distinct and memorable phrases, making them easier for models to recognize when presented as answer choices. In contrast, the PtS task requires the model to recognize a story that matches the proverb's meaning, which demands a deeper interpretation of the story's nuances and the proverb's meaning.

Looking at the medium-level open-sourced models, we can see a big variety of performance. First, we can see the evolution of performance of the Llama Models, LLama1 and Llama2 show performances very close to the random expected score, and the performance improvement from LLama1 to LLama2 is negligible, a possible reason for this is that, while Llama2 trains in the double of tokens, the proportion of Portuguese tokens in it's train is still very low at 0.09\% of the training data~\citep{touvron2023llama2}, the low contact with Portuguese documents may have hindered the models understanding of local expressions. LLama3-8B however, does show an expressive improvement over LLama2-7B, LLama3-8B is trained on 7 times more tokens, but the language distribution of the training data is not known, making it hard to discuss the impact of Portuguese documents in this performance improvement.

Still on the topic of the Llama models, Sabiá-7B is a model further pretrained in 10B Portuguese starting from Llama-7B; we can see that this Portuguese specialization did bring some gains in the StP task, but not so much in the PtS task. It is interesting to see that Sabiá-7B slightly outperforms LLama2-7B by training in Portuguese documents in a much smaller scale, as LLama2 trains in 2T tokens, and Sabiá-7B total train is 1T tokens, herded from using Llama 7B as base, plus 10B Portuguese tokens, this results points to the importance of regional documents in model pretraining.

Looking at the Qwen models, they clearly outperform their counterparts with the same number of parameters and show a progressive improvement as the parameter count goes up, with Qwen 2.5 14B reaching performance close to the commercial models. Qwen models are pretrained in 18T tokens, but little is known about the training data, thus limiting our discussion here. Nevertheless, we can observe a positive correlation between model size and model performance in the BRoverbs tasks.

Finally, looking at the smaller models, Tucano, Tinyllama, and Curió all show performance close to the random expected score of 20\%. This is interesting since we know that a positive score is possible for this model size, since Qwen 1.5B performs well above the random score. However, even with TinyLlama 3T—trained on three times more English tokens than its 1T counterpart—we see no sign of improvement. Moreover, incorporating Portuguese data does not appear to help in this scenario. Curió 1.1B, which builds on TinyLlama 1T with an additional 150B Portuguese tokens, still performs at random chance level. Likewise, Tucano 2.4B, trained from scratch on 500B Portuguese tokens, also fails to show meaningful gains.

\subsection{Does More Portuguese Data Improve Cultural Understanding?}

\begin{figure}
    \centering
    \includegraphics[width=1\linewidth]{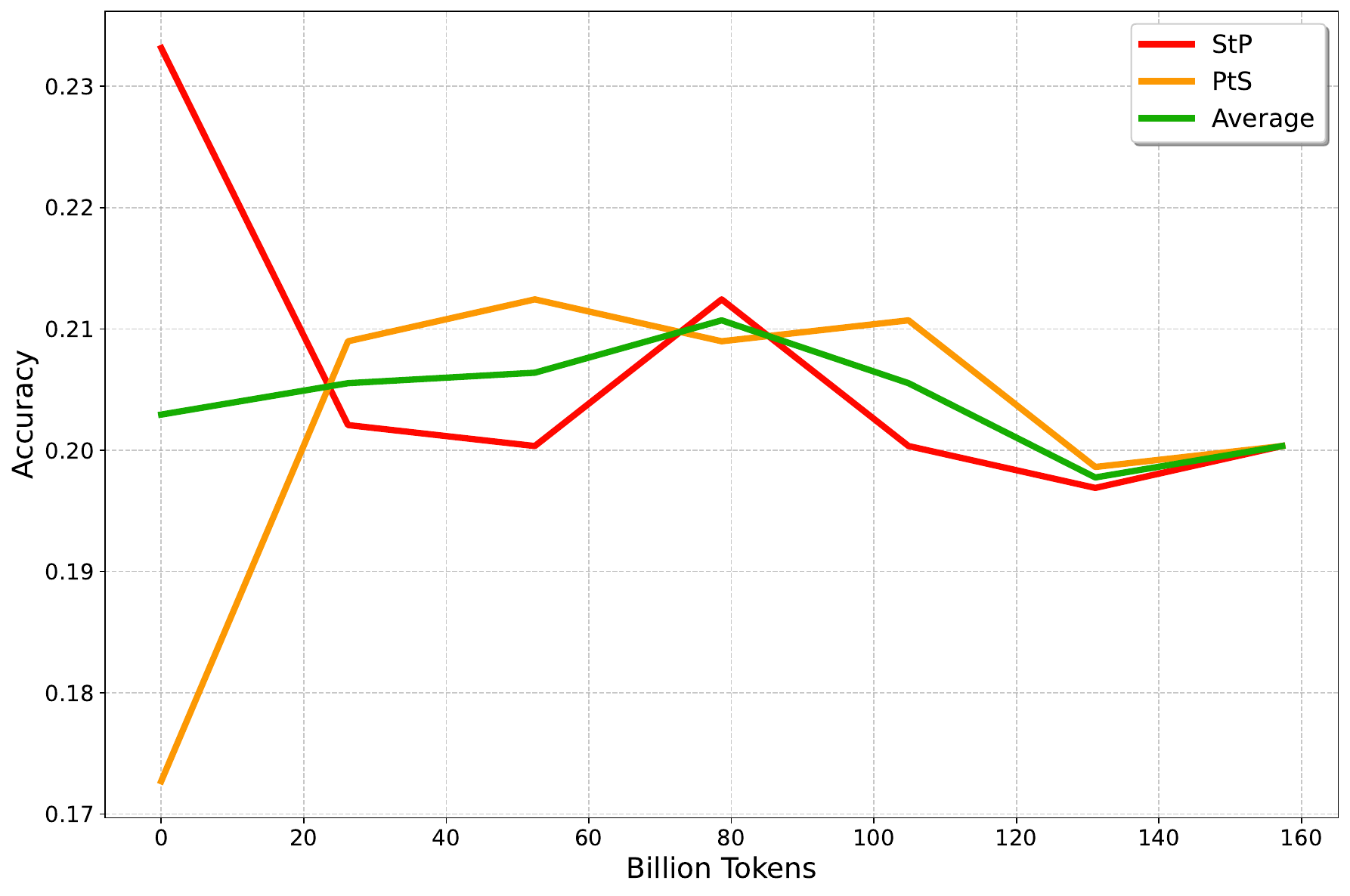}
    \caption{Accuracy of Curió in Broverbs tasks through training}
    \label{fig:broverbs_curio_performance}
\end{figure}

Since intermediate checkpoints for Curió-1.1B are available, we can track their performance throughout training. This setup is particularly relevant because Curió starts from an English-only model and transitions the training to Portuguese data, allowing us to observe the impact of the continued pretraining in our task. Given that our evaluation requires familiarity with Brazilian proverbs, we expected to see performance improvements. Figure~\ref{fig:broverbs_curio_performance} shows Curió-1.1 B's results over time for both BRoverbs tasks.

However, the results are mixed. While PtS shows performance slightly improves, StP declines, leading to no overall gain compared to TinyLlama 1T, as well as the average score never distancing itself from the random performance of 20\% accuracy. This suggests that despite training on Portuguese text, the data may not have included enough relevant information about Brazilian proverbs. Another possibility is that Curió-1.1B requires more training for this ability to emerge, as seen in some large-scale studies of emergent capabilities~\citep{wei2022emergent}. 

The impact of training data quality and language exposure on task performance remains unclear. For instance, TinyLlama 3T, despite being trained on 3 trillion English tokens, performs at chance level, while Qwen 2.5 1.5B achieves notably better results after 18 trillion tokens. However, since details about Qwen's training data are not publicly available, it's difficult to pinpoint the exact cause of this improvement. It could arise from the benefits of large-scale training on general reasoning capabilities, the presence of substantial Portuguese content in Qwen's training corpus, or a combination of both factors.

The results from Sabiá-7B and Curió-1.1B suggest that models may need to reach a certain capacity threshold to handle our task effectively. This threshold doesn't necessarily require pretraining on Portuguese data; however, once the model attains it, exposure to Portuguese—especially documents that are better aligned with the task—can significantly enhance performance.

\section{Future Work}

In our analysis, we use the amount of Portuguese data included in a model’s pretraining as a proxy for its potential exposure to Brazilian proverbs, since such proverbs are more likely to appear in Portuguese documents. A more direct approach for future work would be to measure the actual presence of Brazilian proverbs in large Portuguese corpora, such as Gigaverbo~\citep{correa2024tucano} (used for training Tucano) and the Portuguese subset of Clueweb22~\citep{overwijk2022clueweb22} (used for continual pretraining of Sábia 7B).

Our results also indicate that current state-of-the-art commercial models already achieve strong performance on the current version of the dataset. To increase the difficulty of the BRoverbs benchmark, future iterations could include more than five alternatives per question. We limited the present version to five alternatives to keep manual validation feasible, as verifying a larger set of options for each question becomes increasingly challenging.

While our work introduces a benchmark and resources for evaluating Brazilian proverbs in Portuguese. Future work can use our benchmark, in addition to others such as ProverbEval~\citep{azime2024proverbeval}, to make a more nuanced analysis of LLMs' abilities to understand proverbs across multiple regions and languages.

Finally, the selection of models used in this work is still limited; Future work could expand the types and sizes of models evaluated for both open and closed source models, possibly achieving further insights.

\section{Conclusion}

In this paper, we introduced BRoverbs, a new benchmark designed to evaluate LLMs' ability to understand and associate Brazilian proverbs with corresponding stories. The dataset consists of two core tasks: Proverb to Story, where a model must select the correct story that embodies a given proverb, and Story to Proverb, where a model must identify the proverb that best represents a given story. Through these tasks, BRoverbs provides insights into how well LLMs capture linguistic and cultural nuances in Brazilian Portuguese.

Our evaluation of various LLMs revealed a significant disparity in performance across model sizes and architectures. Large commercial models such as GPT-4o and Sabiá-3 achieved near-perfect scores, while smaller open-source models struggled, with some performing close to random selection. Notably, Qwen models demonstrated superior performance among open-source models.

Our analysis of Curió-1.1B, a model that underwent continued pretraining in Portuguese, revealed no significant gains in BRoverbs performance despite its exposure to 150B additional tokens. Meanwhile, Sabiá-7B, which underwent continued pretraining in 10B, showed slightly improved performance in the StP task. This raises questions about the role of pretraining data composition versus the overall training and model scale. Future work should explore whether specific exposure to idiomatic expressions and cultural narratives could enhance proverb comprehension.

Overall, BRoverbs advances the scenario of culturally grounded evaluation datasets for Brazil. As LLMs continue to evolve, benchmarks like BRoverbs could play a crucial role in ensuring fair and accurate assessments of model capabilities in underrepresented languages.

\section*{Declarations}

\begin{acknowledgements}
We would like to thank Maritaca AI for providing the necessary computational resources for this research.
\end{acknowledgements}

\begin{funding}

The computational resources used in this Research were provided by Maritaca AI.
\end{funding}

\begin{contributions}

We enumerate below the areas of contribution of the three authors: Thales Sales Almeida (TSA), Giovana Kerche Bonas (GKB), and João Guilherme Alves Santos (JGAS). We follow the CRediT taxonomy\footnote{https://credit.niso.org/}

\begin{itemize}

    \item \textbf{Conceptualization} –-- TSA
    \item \textbf{Data curation} –-- JGAS
    \item \textbf{Formal analysis} –-- GKB
    \item \textbf{Funding acquisition} --- TSA
    \item \textbf{Investigation} –-- TSA
    \item \textbf{Methodology} –-- TSA, GKB, JGAS
    \item \textbf{Project administration} –-- TSA
    \item \textbf{Resources} –-- TSA
    \item \textbf{Software} –-- TSA, GKB, JGAS
    \item \textbf{Supervision} –-- TSA
    \item \textbf{Validation} –-- TSA
    \item \textbf{Visualization} –-- TSA, GKB, JGAS
    \item \textbf{Writing –-- original draft} –-- TSA, GKB, JGAS
    \item \textbf{Writing –-- review \& editing} –-- TSA, GKB, JGAS
\end{itemize}

\end{contributions}



\bibliographystyle{apalike-sol}  
\bibliography{references.bib}

\appendix

\section{Appendix}

\subsection{Carbon emission estimates}

As mentioned before, we used an RTX A6000 to perform our evaluations; all experiments ran in under 5 hours. By using the ML CO$_2$ Impact calculator\footnote{\url{https://calculator.linkeddata.es/}}, we arrive at an estimated carbon emission of 0.85 Kg of CO$_2$.

\subsection{Prompts for story generation}
\label{sec:prompts}
This Appendix presents the prompt that was employed to generate the short stories in the dataset. 

\begin{figure}[H]
\centering
\begin{mdframed}
\begin{lstlisting}[breaklines=true]
Você receberá um provérbio brasileiro e deverá criar três 
situações narrativas curtas que ilustrem a ideia desse provérbio.

Regras:
1. A situação deve ser específica e realista, como um pequeno trecho de uma história.
2. Não mencione o provérbio ou palavras-chave dele diretamente. A ideia deve ser transmitida indiretamente.
3. A história deve ter um personagem e uma ação, mostrando uma consequência ou aprendizado.
4. O formato de saída deve ser assim:
{
   "proverbio": "",
   "explicacao": "",
   "historia_curta_1": "",
   "historia_curta_2": "",
   "historia_curta_3": ""
}
"proverbio": repita o mesmo provérbio que recebeu.
"explicacao": explique o sentido ou a moral do provérbio de forma curta.
"historia_curta": crie uma história curta em português com um ou mais personagens
que vivenciem uma situação que ilustre a ideia principal do provérbio. Não cite o
provérbio na história.
Não inclua nada além do JSON de resposta. Por exemplo, não inclua texto fora do
objeto JSON.

Exemplos 
... 

Agora, gere a resposta para este provérbio:
{proverbio}
\end{lstlisting}
\end{mdframed}
\caption{Prompt completo utilizado para geração das histórias (redimensionado).}
\label{fig:prompt}
\end{figure}

\end{document}